# SIP: A Construction-Phase 3D LiDAR Dataset Capturing Real-World Sensing Constraints

Seongyong Kim, S.M. ASCE[1], and Yong Kwon Cho, F. ASCE[2*]

[1]Graduate Student, School of Civil and Environmental Engineering, Georgia Institute of Technology, 790 Atlantic Dr., Atlanta, GA 30332.

[2]Professor, School of Civil and Environmental Engineering, Georgia Institute of Technology, 790 Atlantic Dr., Atlanta, GA 30332. E-mail: yong.cho@ce.gatech.edu (Corresponding Author)

ABSTRACT

LiDAR has become a core sensing modality in construction due to its ability to reliably capture geometric information in cluttered and dynamic environments. However, most publicly available 3D perception datasets are derived from densely fused scans with near-complete visibility, which differs from practical construction data acquisition. On active construction sites, comprehensive scene reconstruction is often neither feasible nor necessary, and field data are therefore commonly collected as isolated single-station LiDAR scans exhibiting radial density decay and fragmented geometry. These conditions motivate perception methods that operate on partial observations without relying on full scene rescanning or global reconstruction. This paper presents SIP (Sites in Pieces), a dataset designed to preserve the realistic sensing characteristics of construction-phase LiDAR while organizing the data within a construction-specific semantic taxonomy. SIP includes indoor and outdoor LiDAR scenes annotated at the point level and categorized into Built Environment, Construction Operations, and Site Surroundings classes. The dataset covers both permanent structural components and temporary objects such as scaffolding, staged materials, and scissor lifts, which often appear sparse or fragmented and pose segmentation challenges. A standardized scanning protocol, annotation workflow, and quality control process ensure dataset consistency, and the dataset is released openly with a companion utility code repository to facilitate integration into modern 3D deep learning pipelines. By retaining real-world sensing artifacts, SIP enables realistic benchmarking for construction-oriented 3D vision and supports automated site understanding and digital-twin applications.

# INTRODUCTION

## Background: 3D Scene Understanding in Construction

3D scene understanding has become a foundational capability in modern construction engineering, driven by the increasing integration of digital technologies into field workflows (Chen et al. 2022). Over the past decade, construction projects have increasingly adopted terrestrial laser scanning (TLS) (Qin et al. 2021; Wang and Cho 2015), photogrammetry (Sestras et al. 2025), and related spatial sensing technologies for applications including progress monitoring (Puri and Turkan 2020), safety assessment, quality assurance and control (Xu et al. 2019), as-built documentation (Sepasgozar et al. 2016), and digital twin development (Liu et al. 2021; Wang et al. 2015).

Among sensing modalities, LiDAR has become prominent because it provides geometry-first measurements that remain reliable in cluttered, partially lit, and dynamically changing jobsite conditions. Unlike camera-based systems, LiDAR captures spatial structure even when visual appearance provides limited information (Zhong et al. 2021). As a result, many construction workflows increasingly depend on LiDAR as their primary source of spatial information, making the ability to interpret raw 3D point clouds essential for automated analysis, robotics, and information modeling in active construction environments (Wu et al. 2021).

Construction environments, however, introduce unique complexities that pose challenges radically different from those seen in conventional curated datasets used in 3D perception research (Chen et al. 2019; Park and Cho 2022). Sites evolve continuously as new materials are installed, temporary assemblies are erected, and equipment and workers move throughout the workspace. The resulting scenes contain a mixture of permanent building elements, temporary construction systems, operational equipment, and miscellaneous site materials, producing geometric and semantic variability rarely represented in existing structured datasets.

Semantic segmentation of point clouds (i.e., assigning class labels to individual points) plays a critical role in making sense of such complex and dynamic environments. Segmentation outputs provide the basis for object detection, progress quantification, hazard identification, robotic navigation, inspection workflows, and automated comparisons between as-designed and as-built models (Nguyen et al. 2020; Zhang et al. 2019; Zhao et al. 2019).

Yet, construction-phase data differ from widely used public 3D datasets in several fundamental ways. First, on-site dynamics introduce persistent occlusion and visual clutter. Equipment, workers, stored materials, and temporary assemblies frequently obstruct the scanner's view, while irregular spatial arrangements create overlapping geometry and missing regions that complicate semantic interpretation. Second, temporal evolution continuously alters the environment. As work progresses, objects appear, disappear, or change configuration, and visibility conditions shift across scans. Because construction spaces seldom follow repetitive spatial patterns, these time-dependent changes limit the generalizability of models trained on more structured datasets. Third, construction scenes often exhibit high intra-class variability and low inter-class variability. Objects within the same category may differ widely in shape or configuration, while components from different categories can share similar geometric characteristics, blurring class boundaries and complicating semantic discrimination.

Taken together, these characteristics underscore the need for datasets that capture authentic construction-site sensing conditions, enabling algorithms that can be implemented reliably in real-world settings.

**Practical Constraint: Single-scan or Partially Registered Field Data**

Despite advances in automated registration, construction sites rarely permit practitioners to capture complete multi-view reconstructions. Field constraints often limit scanning to single scans or small groups of only partially registered scans (Aryan et al. 2021; Kim et al. 2024). A *single scan* is an isolated, single-station LiDAR capture that provides a self-contained but inherently incomplete view of the scene. A *partially registered scan* comprises a small set of neighboring scans that can be locally aligned but still cover only a limited portion of the site.

Scanner placement is often constrained by safety boundaries, obstructed paths, and ongoing site activities, limiting opportunities for multi-view acquisition. Physical obstructions such as partitions, equipment, temporary installations, and accumulated materials further prevent ideal positioning, while tight daily schedules encourage rapid and opportunistic data capture rather than deliberate multi-position planning.

Consequently, LiDAR data collected on real job sites rarely resemble the fully registered and geometrically complete models commonly used in research datasets. Instead, they consistently exhibit:

- *Radiating scan geometry*. Visibility decreases with distance from the scanner, producing radial density decay and increasing regions of missing geometry.
- *Partial surfaces*. Objects are often captured from a single vantage point, leaving unseen regions and producing fragmented geometry
- *Irregular sampling*. Angular sampling concentrates measurements near the scanner while distant or oblique surfaces receive fewer points.
- *Unregistered scenes*. Individual scans remain independent captures rather than parts of globally reconstructed models.

Even in this incomplete state, single-scan measurements still contain meaningful information for construction 3D vision tasks. Interpreting such observations enables direct use of field data without additional acquisition or reconstruction, motivating datasets that reflect practical construction sensing conditions.

**Landscape of Public 3D Datasets for Deep Learning**

While the computer vision and robotics communities have produced several influential 3D datasets, none capture the sensing conditions characteristic of construction-phase scanning. Popular indoor datasets such as S3DIS (Armeni et al. 2016), ScanNet (Dai et al. 2017), and Matterport3D (Chang et al. 2017) provide high-quality multi-view reconstructions of structured indoor environments including offices, homes, and public buildings. These datasets generally reflect stable indoor environments with broad visibility and well-structured spatial layouts.

Outdoor datasets, including SemanticKITTI (Behley et al. 2021), Semantic3D (Hackel et al. 2017), and DALES (Varney et al. 2020) are primarily designed for autonomous driving or aerial mapping applications. Their sensor perspectives and semantic categories reflect roadway environments, including vehicles, streets, and surrounding infrastructure.

Synthetic datasets such as ShapeNet (Chang et al. 2015), Objaverse (Deitke et al. 2023), and other graphics-generated repositories provide fully reconstructed object geometry with idealized sampling and clean boundaries. These datasets are valuable for controlled shape understanding tasks but represent simplified geometric conditions.

Dynamic datasets such as NSS (Sun et al. 2023) capture temporal changes using multi-view RGB imagery and derived depth to reconstruct evolving scenes across multiple phases. These datasets provide comprehensive multi-view visibility and detailed observations of structural evolution.

A comparative review of these datasets (see Table 1) highlights a common assumption: most are collected in stable environments with consistent viewpoints and near-complete geometric coverage. In contrast, construction-phase LiDAR data are typically acquired as isolated scans with limited viewpoints, partial visibility, and irregular sampling patterns.

**Table 1.** Comparison of public 3D datasets used in 3D perception research.

| Dataset | Domain Description | Acquisition Modality | Visibility |
|---|---|---|---|
| *S3DIS* | • Indoor; office-like; minimal clutter | TLS | • Multi-view, complete or near-complete visibility |
| *ScanNet* | • Indoor; consumer/residential; curated | RGB-D | • Multi-view reconstructed; mostly complete with partial occlusion |
| *Matterport3D* | • Indoor; real-estate style environments | RGB-D | • Panoramic registered views; complete visibility |
| *SemanticKITTI* | • Outdoor driving; vehicle-centric | 64-beam LiDAR | • Sequential driving viewpoint; sparse object visibility and partial silhouettes |
| *Semantic3D* | • Outdoor static; urban plazas | TLS | • Fully registered static scans; very complete with minimal occlusion |
| *DALES* | • Aerial; rooftops and terrain | ALS | • Top-down coverage; limited side-geometry visibility |
| *Objaverse* | • Synthetic; complete 3D assets | Synthetic | • Fully visible synthetic geometry; no occlusion |
| *NSS* | • Construction-specific dynamic scenes; multi-phase change capture | Multi-view stereo | • Dynamic temporal changes; Multi-view captures with complete visibility |
| *SIP (Ours)* | • Construction-phase; realistic jobsite scan conditions | TLS | • Single-viewpoint captures; severely occluded, incomplete partial-view |

**Distinctive Contributions of the SIP Dataset**

To bridge this gap, we introduce SIP (Sites in Pieces), a dataset designed to capture construction-phase point clouds in the form in which they are realistically acquired in the field. SIP consists of disaggregated, individual terrestrial LiDAR scans collected using a FARO TLS, preserving the raw characteristics intrinsic to jobsite scanning, including partial visibility, limited viewpoint coverage, and sampling inconsistencies. The dataset is deliberately curated and organized with consistent annotations, metadata, and data structures, enabling reliable reuse and reproducible experimentation across studies.

The dataset encompasses a diverse range of construction elements, covering both permanent components (such as ceilings, floors, columns, and partitions) and temporary or slender objects (including pipes, ladders, guardrails, material piles, and other transient assemblies). All scans are annotated at the point level following a structured taxonomy tailored to construction environments, addressing a gap in existing 3D scene datasets, which typically lack semantic labels representing construction objects and jobsite-specific elements.

The key contributions of SIP can be summarized as follows:

- *A curated construction-phase 3D dataset capturing realistic single-scan sensing conditions,* preserving occlusions, partial visibility, and density imbalance typical of jobsite LiDAR observation across indoor and outdoor construction settings.
- *A construction-specific semantic annotation scheme,* including both permanent building elements and temporary construction components that are often absent from existing datasets.
- *A structured and reusable dataset resource,* with consistent annotations, metadata, and supporting utilities that facilitate reproducible experimentation and integration into modern 3D perception pipelines.
- *A benchmark resource for construction-oriented 3D perception,* enabling the development and evaluation of algorithms designed to operate under incomplete observations typical of real construction data.

Collectively, these contributions enable researchers and practitioners to develop and evaluate 3D perception methods that operate robustly under the genuine constraints of active construction sites.

### Intended Use and Scope

The SIP dataset is intended to support the development and evaluation of construction-oriented 3D perception algorithms, particularly point-cloud semantic segmentation and related scene understanding tasks under partial observation conditions typical of construction sites. Because each scan represents a discrete observation of a construction state, SIP enables incremental updating of as-is site conditions directly from single scans without requiring full scene rescanning or global reconstruction. For example, a semantic segmentation model trained on SIP can classify structural and operational elements from a newly acquired LiDAR scan of an active jobsite, allowing localized updates to existing as-is representations, progress tracking, or detection of newly installed components.

SIP is particularly suited for research on construction-phase scene understanding under incomplete or occluded observations. The dataset also supports benchmarking of learning strategies designed for irregular point densities and single-scan sensing conditions. However, SIP is not intended for applications requiring globally registered multi-scan reconstructions, high-fidelity geometric modeling of complete structures, or tasks relying on temporally synchronized multi-view observations.

The SIP dataset is organized in alignment with the Findable, Accessible, Interoperable, and Reusable (FAIR) principles (Wilkinson et al. 2016) to support transparent sharing and long-term use. The dataset is made findable and openly accessible through a persistent public repository and accompanying documentation. Its structure and metadata are designed to promote interoperability with common data-processing environments, while clear licensing and documentation facilitate reuse by researchers, practitioners, and educators.

## DATASET DESCRIPTION

### Dataset Philosophy and Acquisition Context

The SIP dataset emphasizes single-scan realism to reflect how LiDAR is typically collected on active construction sites. Rather than reconstructing multi-scan scenes, each sample corresponds to an individual TLS observation that captures only the geometry visible from a single scanner position. This design preserves partial visibility, occlusions, and radial density gradients inherent to field-acquired data. Consequently, SIP represents the incomplete and view-dependent observations that 3D segmentation models must interpret in practice, where object visibility depends strongly on scanner placement and ongoing construction activities (Figure 1).

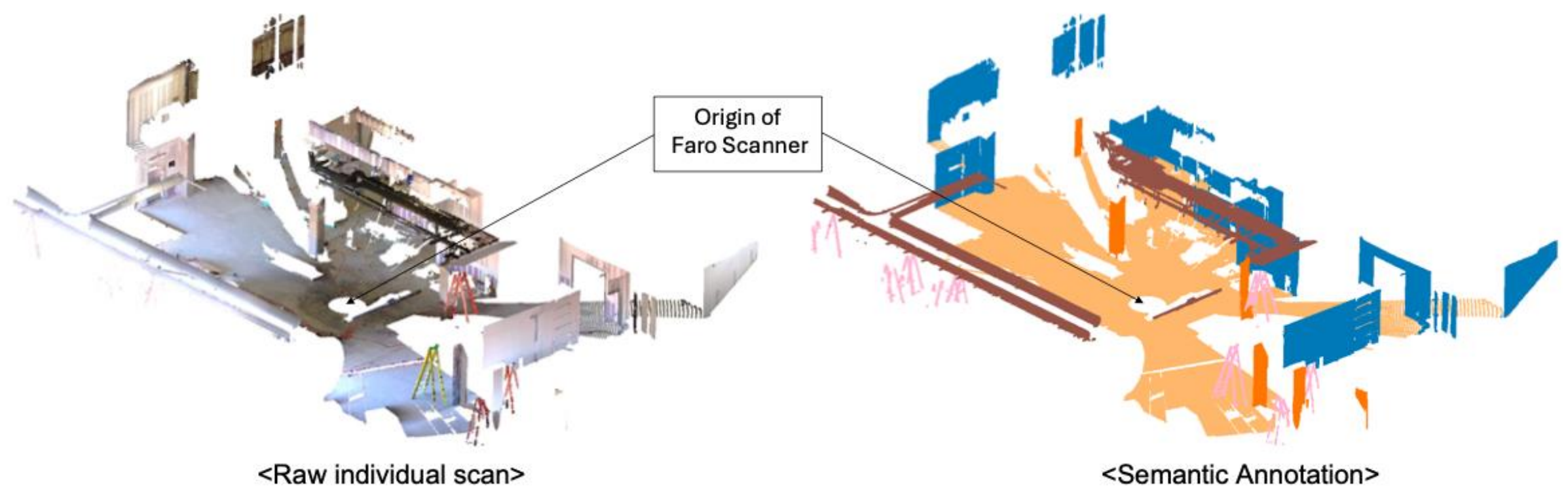

**Fig. 1.** Illustration from the SIP dataset demonstrating single-scan LiDAR acquisition. The raw scan (left) and semantic annotation (right) highlight clutter, occlusion, the scanner origin, and the radial spread pattern of outward-propagating LiDAR returns.

### Composition of SIP Scenes and Dataset

*Dataset Overview*

SIP captures diverse indoor and outdoor construction environments through isolated single-station LiDAR acquisitions. The scenes reflect typical jobsite conditions, including partially completed structures, exposed building systems, temporary equipment, and irregular material layouts (Figure 2). In addition to construction-phase complexity, the dataset preserves sensing characteristics inherent to terrestrial LiDAR acquisition. As illustrated in Figure 3, SIP scans exhibit a pronounced center-weighted point density caused by the radial sampling pattern of single-station measurements. Combined with structural occlusions and constraints on scanner placement, these effects produce point clouds with uneven sampling and fragmented geometry that differ from the more uniformly reconstructed scenes aggregated from multiple viewpoints, as found in datasets such as S3DIS.

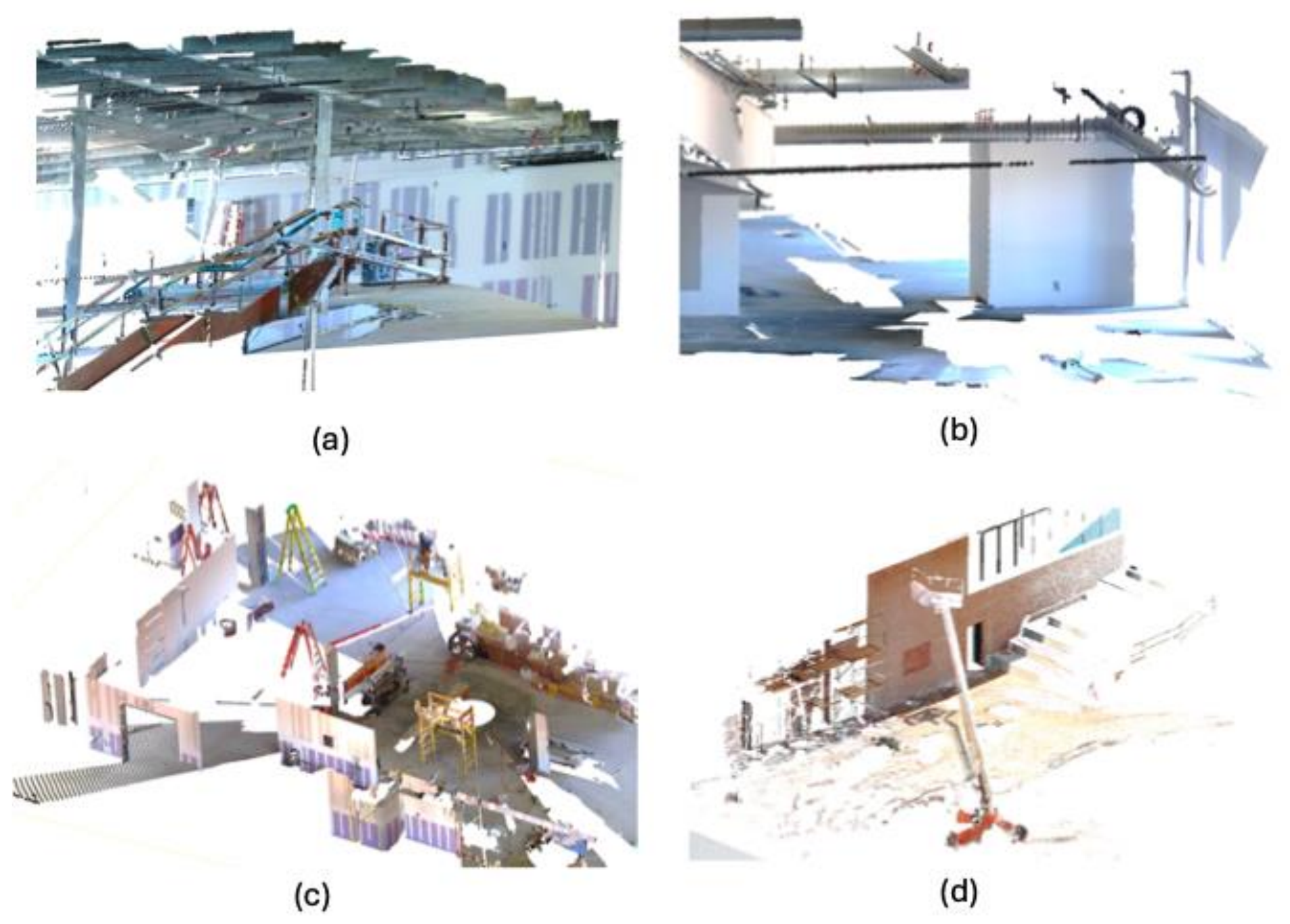

**Fig. 2.** Representative construction-phase elements captured in SIP scans: (a) stair structures in the midst of framing, (b) exposed mechanical and piping systems prior to enclosure, (c) temporary construction equipment including ladders, lifts, and materials, and (d) scaffolding and partially completed exterior walls.

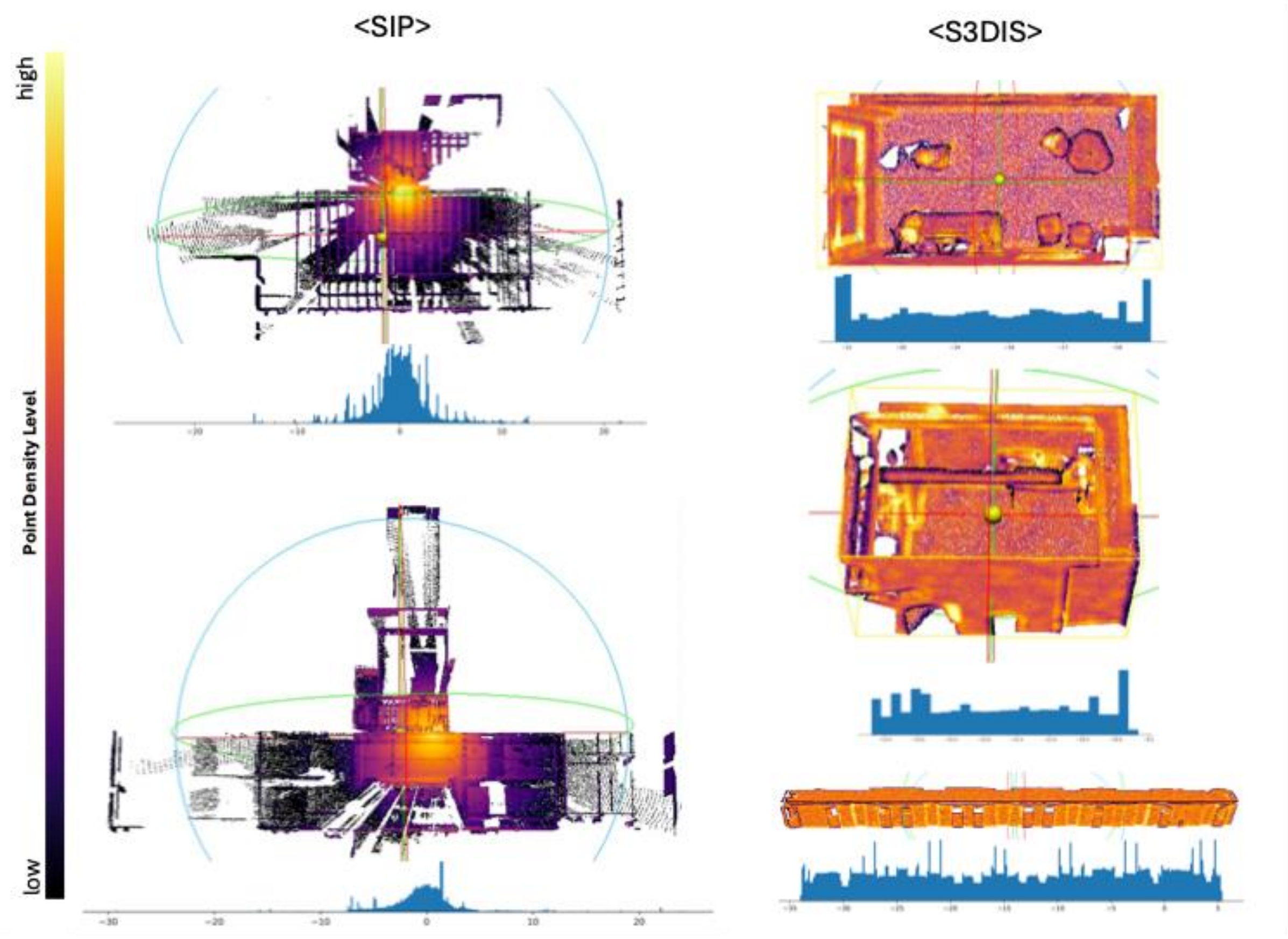

**Fig. 3.** Point-density comparison between SIP and S3DIS scenes. Density computed within a 0.15 m radius; histogram bin size = 0.15 m. All scales are relative.

SIP consists of 40 single-scan LiDAR scenes, comprising 27 indoor and 13 outdoor construction environments, and provides 23 semantic classes spanning permanent structural elements, temporary construction objects, and site-context features. The dataset spans a wide range of spatial footprints, from compact rooms to large open interiors and exterior site environments, leading to substantial variation in point counts across scenes. Each indoor scan averages approximately 3 million points, while each outdoor scan averages approximately 5 million points, reflecting the broader, more variable coverage typical of exterior construction environments (Figure 4). These variations in scale, density, and scene structure provide a meaningful basis for evaluating how segmentation and perception methods generalize across contrasting construction conditions.

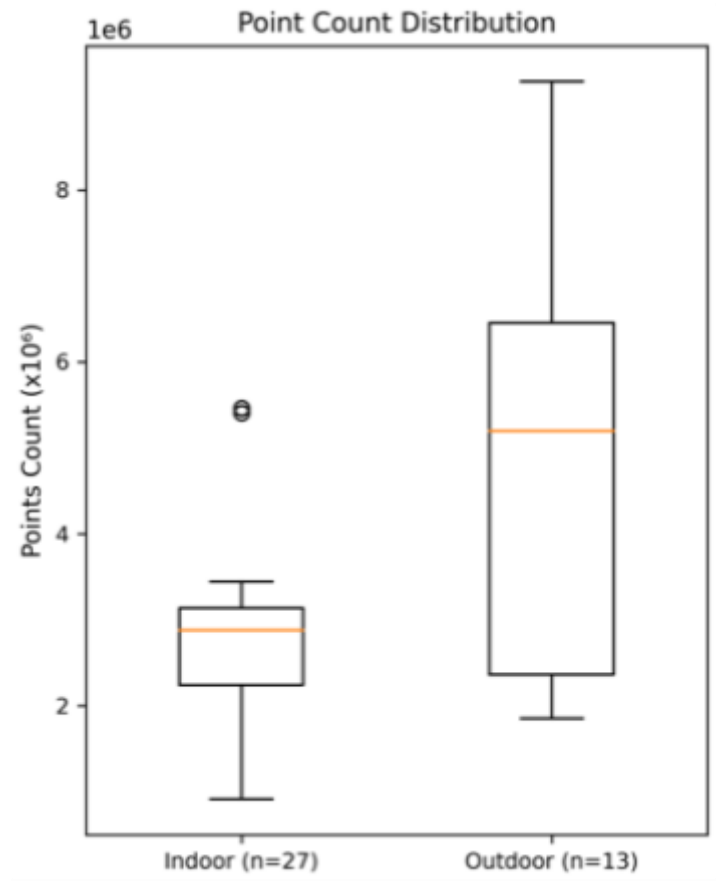

**Fig. 4.** Per-scan point count distribution for indoor and outdoor SIP scenes, showing the larger variation and higher overall point counts present in outdoor scans.

*Indoor Scene Composition*

Indoor scenes include open halls, enclosed rooms, corridors, and stairwells (Table 2). These spaces represent common interior conditions encountered during construction scanning. Open halls offer wide visibility and expansive floor areas; enclosures form tight bounded volumes; corridors create narrow elongated pathways; and stairwells introduce vertical circulation with inherent occlusion, as shown in Figure 5. The paired red-green-blue (RGB) and semantic views illustrate both the spatial composition of these environments and the annotation structure used in the dataset. These spatial patterns illustrate the diversity of interior settings encountered during scanning and provide contextual understanding of the indoor scenes represented in the dataset.

**Table 2.** Indoor SIP scan distribution by spatial pattern and variant.

| Category (with variant) | Scan count | Subtotal |
| --- | --- | --- |
| Open Hall | — | 20 |
| Core | 15 | |
| w/ staircase | 3 | |
| w/ enclosure | 1 | |
| w/ corridor | 1 | |
| Enclosure | — | 5 |
| Core | 3 | |
| w/ corridor | 2 | |
| Corridor | 1 | 1 |
| Stairwell | 1 | 1 |
| | | 27 |

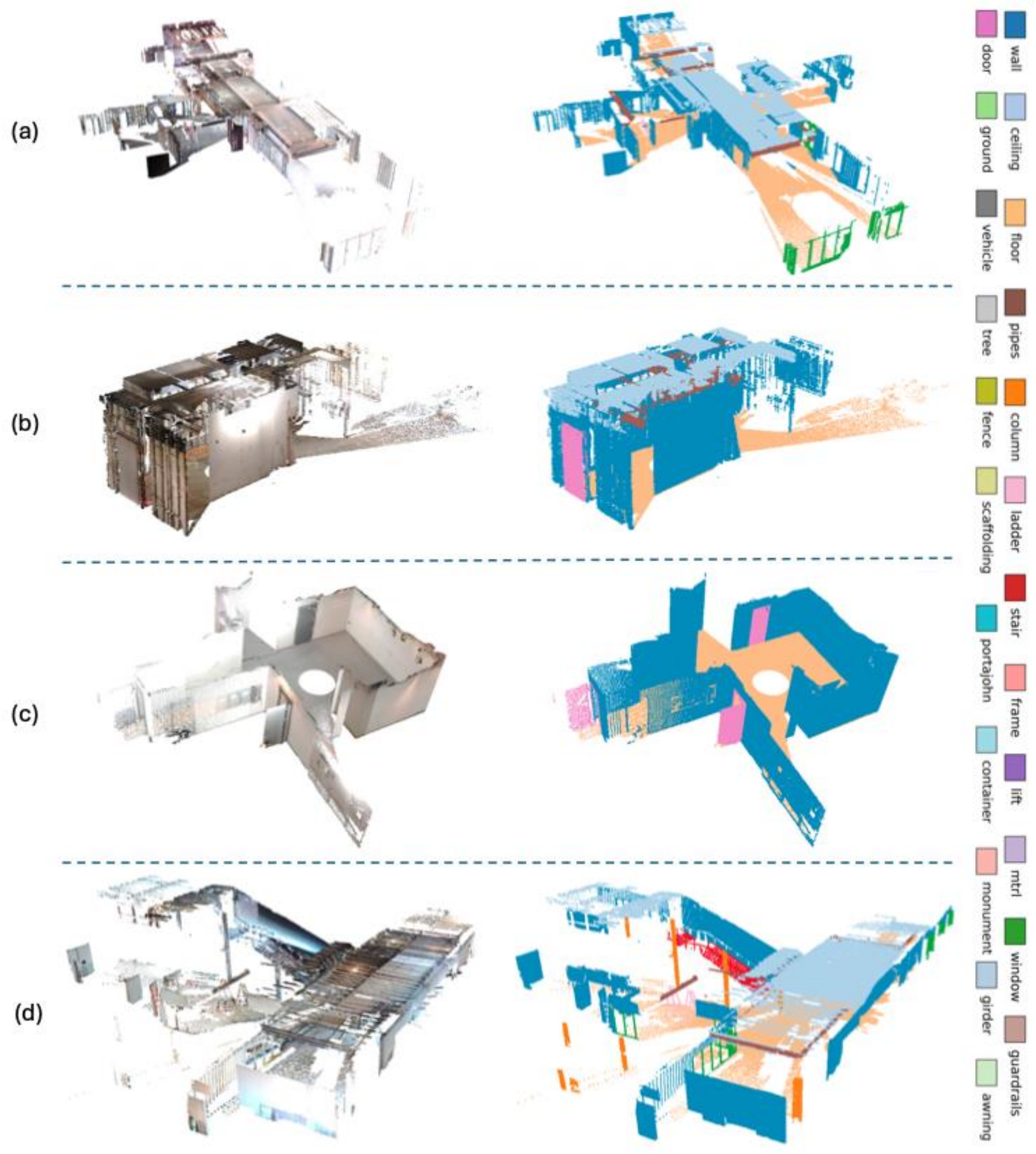

**Fig. 5.** Representative indoor scan examples from the SIP dataset. RGB scans (left) and corresponding semantic annotations (right) are shown for four spatial configurations: (a) open halls, (b) enclosed rooms, (c) corridors, and (d) stair areas. Colors follow the SIP semantic legend at right.

Across these interiors, partially completed structural elements, such as walls, ceilings, floors, and columns, form the primary geometry, while exposed mechanical, electrical, and plumbing (MEP) systems such as ducts, conduits, and piping reflect ongoing construction work. Temporary items, including ladders, scissor lifts, opening frames, scaffolding components, and stored materials, frequently appear in the scans, introducing clutter and irregular object arrangements.

Indoor scanning conditions also shape the data. Shorter distances between the scanner and surrounding surfaces produce higher point density and more complete geometry than outdoor environments. Because indoor spaces are often bounded by ceilings and surrounding walls, a larger portion of the full 360-degree view is captured as meaningful geometry, with fewer empty regions and less wasted scan coverage.

*Outdoor Scene Composition*

Outdoor scenes in SIP capture the broader and less structured settings of active construction sites. These scans encompass open work areas, site perimeters, staging zones, and exterior building surroundings, each shaped by variable terrain, larger spatial extents, and constantly changing operational conditions. A representative sample of outdoor scans is shown in Figure 6.

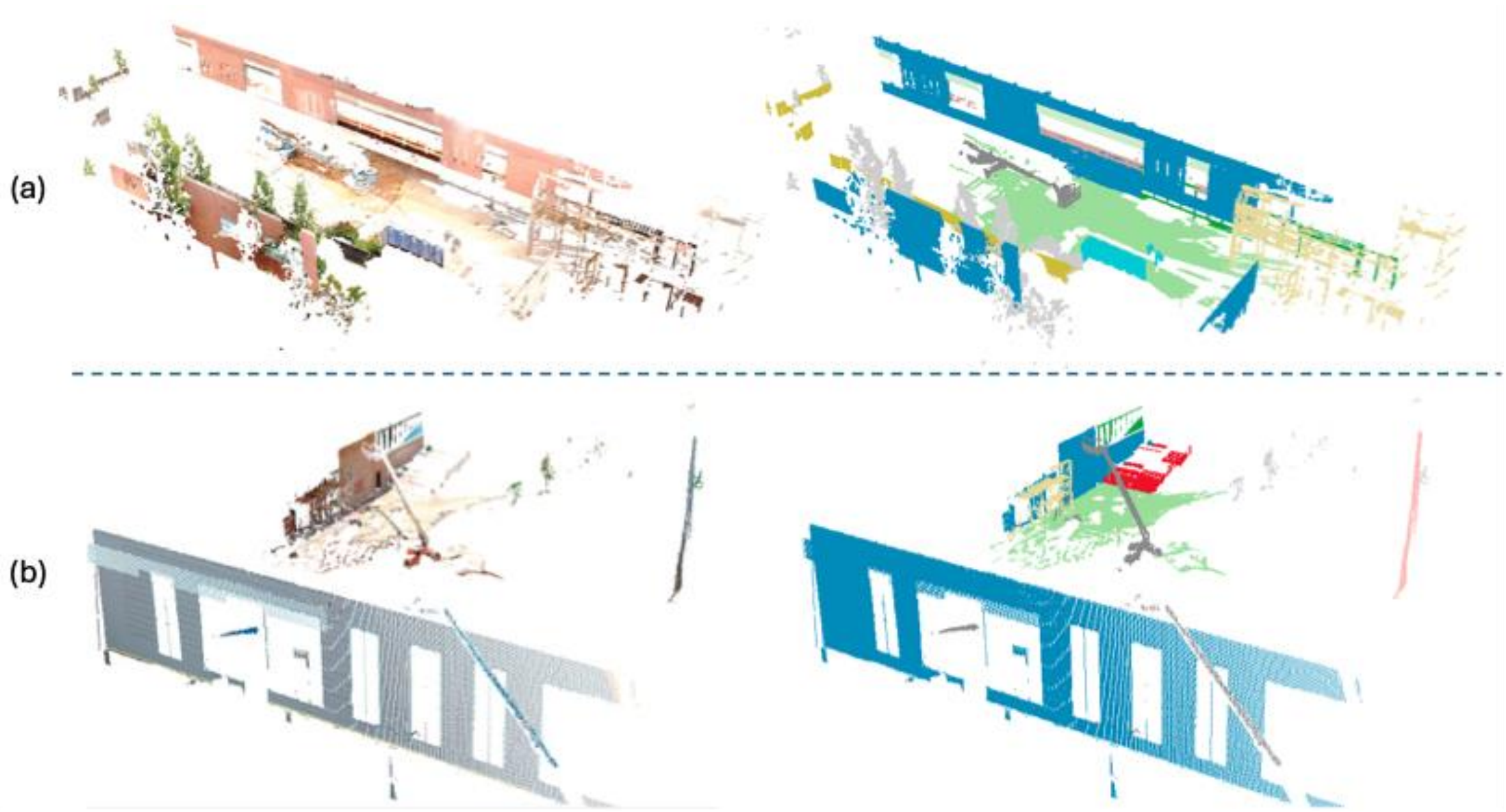

**Fig. 6.** Representative outdoor scans showing exterior construction elements, including ground surfaces, vehicles, fencing, scaffolding, and temporary site infrastructure: (a) façade view; (b) similar exterior view with additional site equipment. RGB scans (left) and semantic annotations (right)

Unlike indoor environments, outdoor scenes lack the enclosing boundaries that constrain visibility and geometry. As a result, they exhibit wide, sparse point distributions and substantial variability in coverage. Ground surfaces, vegetation, temporary fencing, vehicles, and site logistics elements commonly appear, reflecting the diverse exterior conditions encountered during construction.

Outdoor scans also present more complex sensing challenges. Heavy equipment, stored materials, and site layout constraints introduce irregular occlusions, while dynamic activities, such as moving workers, vehicles, or machinery, can produce transient artifacts. Natural lighting, weather exposure, and surface reflectivity variation further influence the appearance and distribution of LiDAR returns.

**Point Cloud Attribute Schema**

Each SIP scan is stored as a point cloud in a simple, platform-independent text format, with each point containing 1) spatial coordinates, 2) RGB color values, 3) LiDAR return intensity, and 4) a surface normal. This straightforward structure allows the data to be easily parsed, visualized, and used for algorithm development without reliance on proprietary formats.

The spatial coordinates (XYZ) represent the 3D position of each point in meters and form the geometric backbone of the scans. Color values (RGB), stored as integers in the 0 to 255 range, provide appearance cues that help distinguish materials and surface conditions in construction environments. Intensity values are stored as integers in the 0 to 255 range and indicate the relative strength of each LiDAR return. FARO typically provides relative intensity values, rather than a calibrated radiometric measurement, and when exported in ASCII format these values fall within the same 0 to 255 range used in SIP. Surface normals in dimensionless unit vectors capture local planar structure and orientation.

Because raw terrestrial LiDAR scans contain millions of points, SIP applies a controlled down-sampling step to make the data more suitable for 3D model training workflows. After point-level semantic annotation, each scan is down-sampled using uniform random sampling, which preserves the natural radial density decay and view-dependent sampling imbalance characteristic of single-station LiDAR acquisition. This approach was preferred over voxel-based down-sampling, which tends to spatially homogenize point distributions and suppress the density gradients inherent to real scanning conditions. By retaining these patterns, the dataset maintains sensing characteristics that are representative of field-acquired LiDAR data. The sampling ratio is determined individually for each scan to account for variations in spatial coverage and point density across different site layouts and scanner placements.

For each scan, the sampling ratio is selected such that the resulting point count is comparable to voxel-based down-sampling at approximately 0.01m, a deliberately conservative reference rather than a strict spacing constraint. This design maintains compatibility with the effective point resolutions (0.02–0.05 m) commonly adopted in public 3D segmentation datasets (Armeni et al. 2016; Chang et al. 2017; Dai et al. 2017), while allowing researchers to apply alternative sampling if desired.

**Class Taxonomy**

SIP defines 23 classes grouped into three functional categories that reflect their roles in construction settings. This taxonomy provides a clear and simple structure for interpreting objects across indoor and outdoor scenes while capturing the diversity of elements present in construction-phase environments (Table 3).

A. *Built Environment Elements:* Permanent or in-progress components of the built structure.

   A1. Permanent Structural Components- 1) wall, 2) ceiling, 3) floor, 4) column, 5) stair framing, 6) girder

   A2. In-Progress Built Components- 7) MEP piping, 8) opening framing, 9) window, 10) door, 11) awning

B. *Construction Operations Elements:* Elements that support construction activities, on-site operations, and the handling or movement of equipment and materials.

   B1. Works & Access Structures- 12) guardrails, 13) site fence, 14) scaffolding

   B2. Operational Equipment- 15) step ladder, 16) scissor lift, 17) construction vehicle

   B3. Site Logistics- 18) staged materials, 19) porta john, 20) office trailer

C. *Site Surroundings:* Natural or site-context elements that are not part of the constructed facility

   - 21) ground, 22) tree, 23) monument

**Table 3**. SIP class applicability for indoor and outdoor scenes, organized by functional group

| Class Group | Indoor | Outdoor |
|---|---|---|
| A. Built Environment | | |
| A1. Permanent Structural | wall, ceiling, floor, column, stair framing, girder | wall, floor, girder, awning |
| A2. In-Progress Built | MEP piping, window, door, opening framing | window |
| B. Construction Operations | | |
| B1. Access Structures | guardrails, site fence | guardrails, site fence, scaffolding |
| B2. Operational Equipment | step ladder, scissor lift | construction vehicle |
| B3. Site Logistics | staged materials | staged materials, office trailer, porta john |
| C. Site Surroundings | — | ground, tree, monument |

# METHODOLOGY AND DATA COLLECTION

## Dataset Collection Protocol

All scans were collected using the FARO terrestrial laser scanner equipped with a dual-capture system that simultaneously records LiDAR geometry and 70-megapixel color imagery. The sensor provides a 300° vertical and full 360° horizontal field of view, using a Class 1 laser with a divergence of 0.19 mrad (0.011°). The angular sampling step is approximately 0.009° in both axes, and the manufacturer-specified ranging error is ±0.2 mm when noise compression is disabled. Each scan required roughly 15 minutes to complete, depending on resolution settings, including concurrent RGB capture. Outdoor data collection was conducted under clear weather conditions to maintain consistent surface visibility and stable LiDAR return characteristics during scanning. The scanner was mounted on a tripod at a consistent height of approximately 1.4–1.5 m for all scans.

Although scan settings, spatial coverage, and effective sensing distance inevitably vary depending on site conditions and accessibility, ensuring an effective point resolution of approximately 0.01 m or finer is sufficient to obtain 3D-model-trainable data under comparable conditions. Consistency in data distribution across scenes is therefore achieved through a standardized post-acquisition down-sampling procedure that normalizes point density rather than through identical acquisition settings.

Data were collected at the student center construction project on the Georgia Tech campus (Georgia Institute of Technology 2025). The interior spaces were undergoing phased construction across three floors, while the exterior frontage hosted ongoing work involving equipment, materials, and active site traffic (Figure 7). A total of 40 scans were collected over a 5-month period during interior finishing operations, including drywall installation, pipe routing, mechanical work, and stair assembly, as well as selected exterior activities.

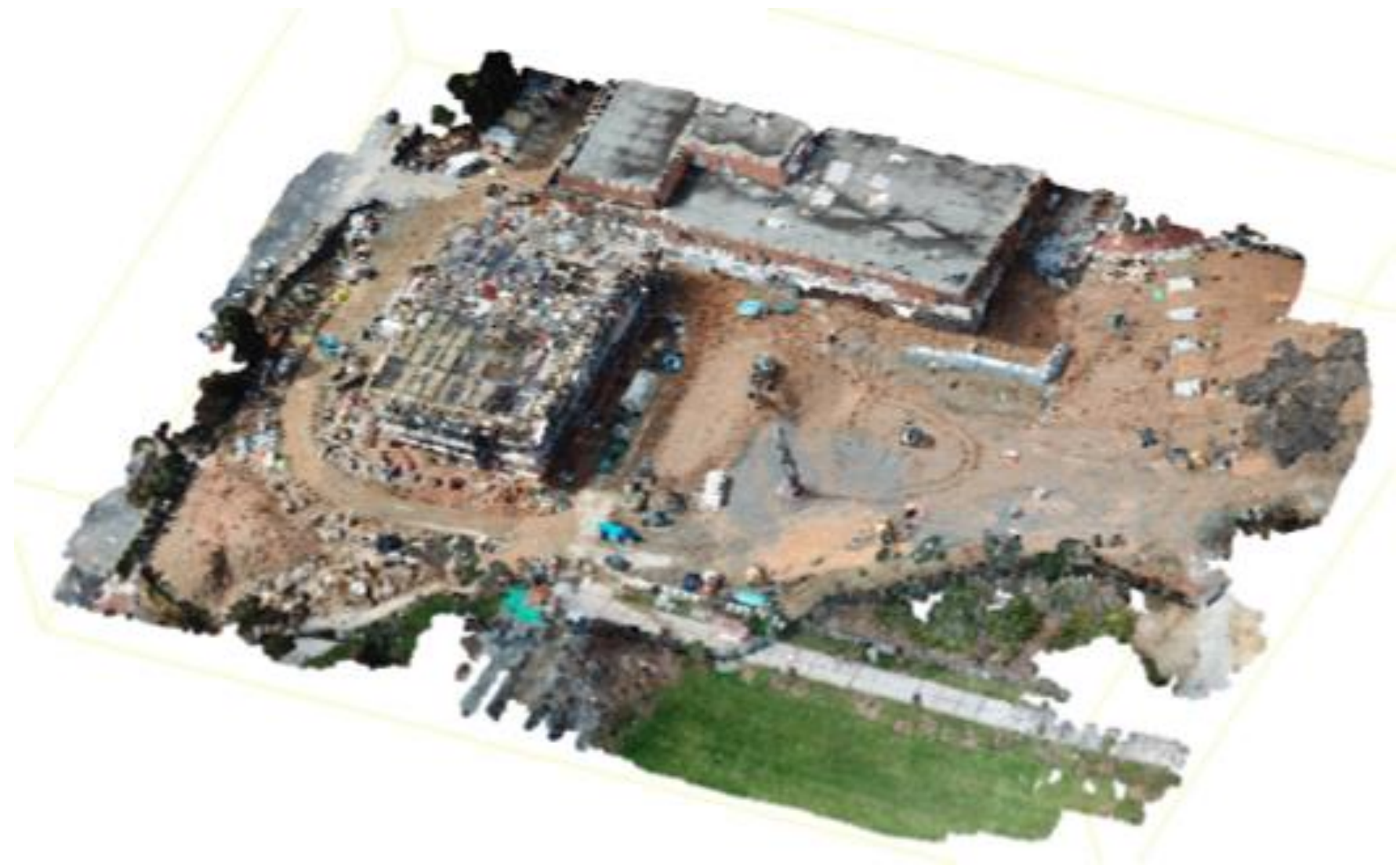

**Fig. 7.** Aerial 3D reconstruction of the John Lewis Student Center site, including the active construction frontage, produced using drone-based photogrammetry.

Scan positions were intentionally selected for single-scan use rather than multi-scan registration. Each vantage point was chosen to maximize coverage from a single observation position, typically the center of a room, open lobby, or strategic corridor intersection (Figure 8). This approach preserved natural visibility limitations and resulted in scenes that realistically reflect practical visibility limitations. Indoor scenes generally provided clearer line-of-sight conditions, while outdoor scenes involved larger standoff distances, façade-level geometry, and increased variability. No registration or global alignment was performed so that each scan remained an independent representation of a construction state.

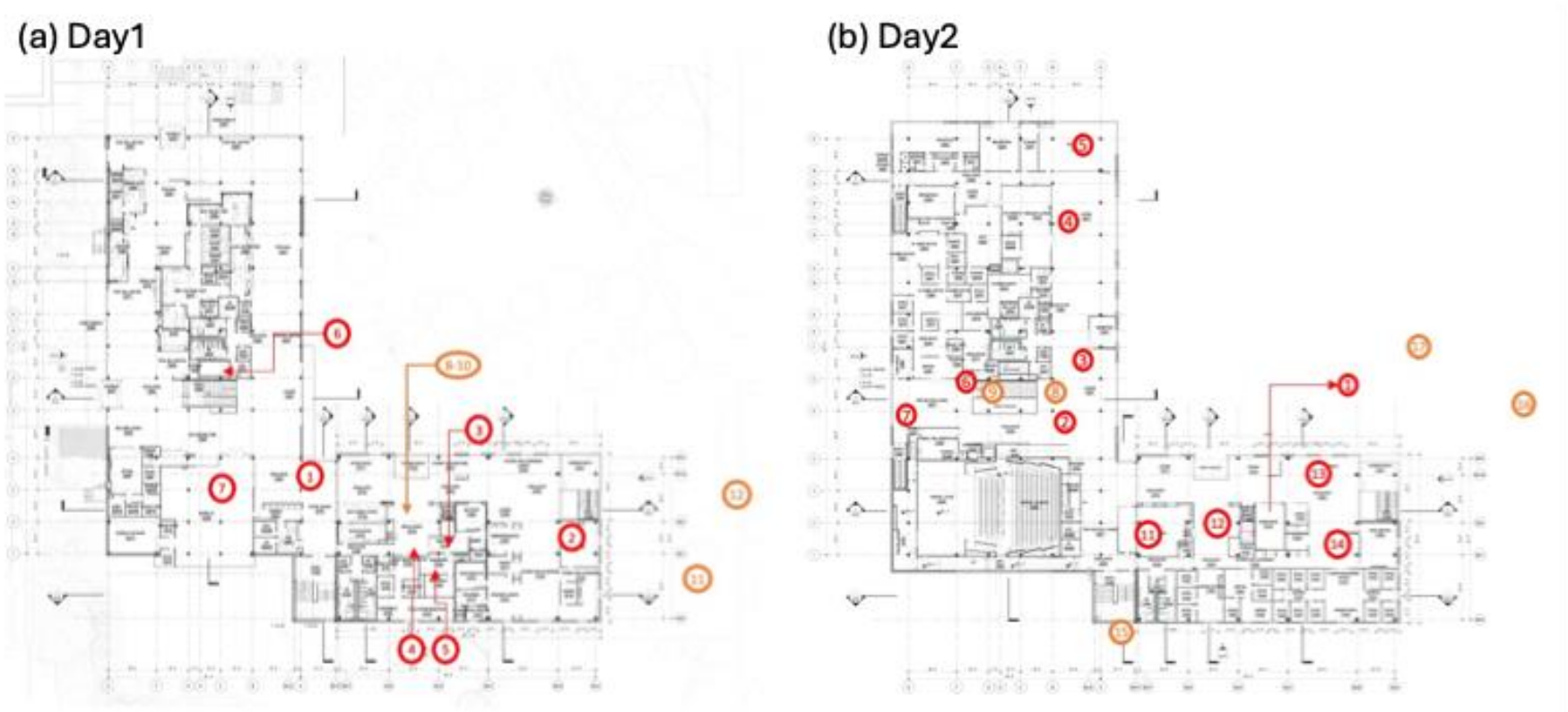

**Fig. 8.** Terrestrial LiDAR scanning positions overlaid on floor plans of the Georgia Tech Student Center construction site. Panels show scan placement from two representative acquisition days.

Site access was constrained by active construction operations, limiting the areas and time windows available for scanning. Data collection therefore occurred during short periods, typically around lunch hours, to minimize interference with site activities. Scanning position choices were guided by practical constraints, safety boundaries, obstructed paths, and worker movement near the scanner. Data collection was therefore carried out during short time windows, typically around lunch hours to reduce interference with site operations. Even under these condition, the environment remained dynamic due to constant worker activity and continually shifting site conditions.

**Annotation Procedure**

Raw point measurements were collected in the proprietary FARO.fls format, which was later converted into plain ASCII .txt files containing XYZ coordinates, RGB values, intensity, and surface normals. Normals were computed using the Open3D library when they were not present in the exported files. No smoothing, or filtering beyond basic noise removal was applied in order to preserve the raw characteristics of single-scan LiDAR geometry.

CloudCompare (CloudCompare 2025) was used as an annotation tool. Annotators used simple editing utilities such as rectangular and polygonal selection, along with connectivity-based extraction for object-specific segmentation. No automated or color-based segmentation modules were used to ensure that all labels reflect deliberate human interpretation.

Annotation began with manual removal of isolated noise, ghost points, and small clusters that did not represent physical surfaces. After denoising, annotators segmented each class individually, resulting in per-class files aligned with the SIP dataset structure. Only visible geometry was labeled; occluded surfaces were not inferred or completed.

Three annotators with prior construction field experience, including field-site exposure through general contractors, performed the labeling using a shared guideline. Annotators received brief training on the required 3D editing tools, and all labeled scans were cross-checked before inclusion in the final dataset to ensure consistency with the defined taxonomy.

## DATA VALIDATION AND QUALITY CONTROL

### Validation Rationale

This section summarizes the validation of the SIP dataset with respect to spatial plausibility, annotation correctness, and consistent formatting. Because SIP is composed of single vantage-point scans, validation focuses on internal coherence rather than global completeness. Single-scan construction data inherently include occlusion, uneven density, and limited visibility, making completeness-based validation infeasible. Validation instead ensures that each scan plausibly reflects the visible scene, maintains geometric coherence, and preserves natural field conditions without over-cleaning or reconstructing unseen regions.

The validation workflow progresses from raw scan export to annotation checks, scene-level plausibility review, and final dataset-wide consistency enforcement. Automated scripts detect missing attributes, malformed files, or unusually sparse classes, ensuring uniform standards across all scans.

### Annotation and Scene-Level Quality Control

Annotation quality control combined iterative manual review with structured verification by annotators with construction-domain knowledge. Because consistent labeling of construction scenes requires interpretation of object function and site context, annotations were reviewed to ensure consistent application of class definitions and realistic object boundaries. In regions with sparse observations or limited visibility, labels were assigned only when local geometric context allowed reliable identification; otherwise, observations lacking sufficient evidence were treated as noise. The dataset was also inspected to ensure suitability for machine-learning workflows, including checks on class granularity, treatment of partially visible geometry, and removal of points better interpreted as noise.

Scene-level validation ensured that annotations remained geometrically and contextually plausible. Labels were verified against the cleaned scan to confirm correspondence with visible surfaces, and overlapping or duplicate labels were corrected. Additional checks examined whether architectural elements exhibited expected geometric behavior, such as planar walls, consistent floor elevations, and vertical structural components. Color–geometry alignment was also inspected to detect potential RGB misregistration, and major visible components were confirmed to be annotated without omission.

Despite these procedures, certain ambiguities remain inherent to construction-phase annotation. Construction scenes often exhibit high intra-class variability and partial visibility, which complicate stable class definitions across scans. The single-station acquisition also produces strong point-density decay and limited visibility in distant regions, making slender or temporary elements more difficult to distinguish. As a result, objects such as scaffolding or ladders are typically identifiable only near the scanner, while distant regions more reliably preserve larger structural elements. Although the moderate dataset size in this work allowed expert annotators to maintain labeling consistency, scaling such domain-informed annotation remains a broader challenge for construction 3D datasets.

**Dataset-Wide Consistency Checks**

Dataset-wide checks enforced consistent directory structure, class mapping, and annotation formatting across all indoor and outdoor scans. Automated scripts verified that each scene contained the complete set of point attributes, such as, XYZ coordinates, RGB color values, LiDAR intensity, and surface normals, and that all numeric ranges were physically valid. Additional integrity checks detected malformed files, missing attributes, or annotation folders containing empty or unintentionally sparse classes.

To ensure privacy and uniformity, all scan folders were anonymized using UUID-based identifiers during final packaging, replacing any site- or time-specific naming inherited from the field acquisition phase. Cross-scene consistency was also reviewed by comparing class-frequency patterns to identify outliers or mislabeled categories. Note that no human-subject data or personally identifiable information is included in the dataset. Scenes occasionally contain workers or moving objects, but LiDAR returns are too sparse to reveal identity, and no RGB imagery of people was retained.

A representative figure showing the original, intensity, normal, and annotation panels is included to support misalignment inspection (Figure 9). Viewing these channels side-by-side helps reveal issues such as RGB–geometry drift, inconsistencies in normal orientation, or annotation boundaries that deviate from geometric or radiometric cues. This visual comparison provides a direct, intuitive way to confirm annotation correctness and overall file integrity.

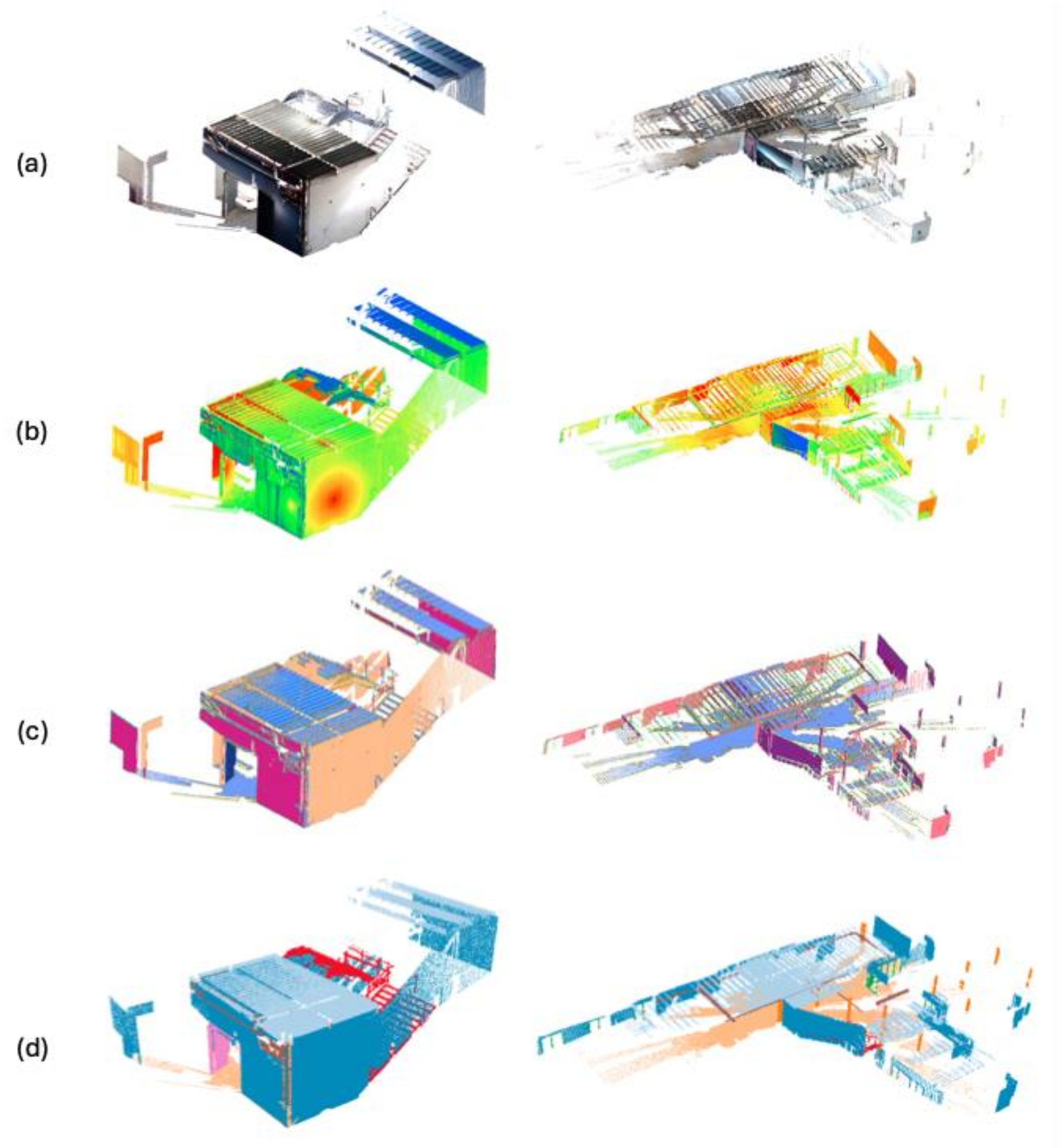

**Fig. 9.** Multi-attribute visualization for misalignment and annotation verification. (a) RGB rendering, (b) LiDAR intensity, (c) normal orientation, and (d) final semantic labels.

## USAGE NOTES

### Dataset Access and Structure

The SIP dataset is openly accessible through its Digital Object Identifier (DOI): https://doi.org/10.5281/zenodo.17667735. Utility scripts and usage instructions are maintained in the accompanying GitHub repository (https://github.com/syoi92/SIP_dataset), with a stable version archived under the dataset DOI. The dataset is released under the CC BY-NC 4.0, while the code is distributed under the MIT license to support transparent reuse and adaptation.

```
sip-indoor/
  | class_config.json
  | splits.json
  └scans/
    | scan.txt [xyzrgbINor] - original scan (pre-annotation)
    └Annotation/
        | class1.txt [xyzrgbINor]
        | class2.txt
        | ⋮
        └classN.txt
```

**Fig. 10.** Directory structure of the SIP-Indoor dataset

Each scan folder follows a consistent directory structure, as illustrated in Figure 10. The '*scans/*' directory contains the down-sampled single-scan point clouds in ASCII .txt format, whereas the '*Annotation/*' directory stores per-class annotation files corresponding to the same scan. The file 'class_config.json' contains class labels, associated RGB color codes for visualization, and an index flag indicating which classes are intended for training or evaluation. Dataset partitions are provided in 'splits.json', which maps each scan to the train, validation, or test split.

Two helper scripts are included for practical use.

- view_anno.py produces class-wise color-coded point clouds by combining the raw scan with its annotation files according to the color definitions in class_config.json. This enables quick verification of annotation quality and scene composition using any standard point-cloud visualization tools.
- preprocessing.py prepares training datasets by reading the selected class indices from class_config.json and dataset partitions from splits.json. It generates train, validation, and test sets containing only the chosen classes and outputs dictionary-style tensors including coordinates, RGB values, intensity, normals, and class labels. To preserve realistic acquisition characteristics, preprocessing also reorders points to approximate the original LiDAR scanning pattern.

**Model Training Pipeline**

The dataset is compatible with contemporary 3D perception models such as PointCept (Pointcept Contributors 2023), MinkowskiEngine (Choy et al. 2019), Point Transformer (Wu et al. 2022, 2024; Zhao et al. 2021), and sparse convolutional backbones (Çiçek et al. 2016). The accompanying repository also provides configuration guidance for applying PointCept-style training pipelines to the SIP dataset.

Users may flexibly configure which classes are included in training through the class configuration file, and the preprocessing script applies this selection to generate datasets aligned with the chosen taxonomy. A typical workflow is as follows:

1. Inspect scans and annotations. View the raw point clouds (scans/) together with the corresponding annotation files (Annotation/) using a 3D visualization tool.
2. Configure training classes. Update class_config.json to specify indexed classes for training. Classes excluded from training are retained as a separate mask (e.g., labeled -1) during preprocessing.
3. Run preprocessing. Execute preprocessing.py, which merges per-class annotations, applies the class indices, and generates train and test subsets based on splits.json.
4. Export to framework format. The resulting tensors (coordinates, RGB, intensity, normals, labels) can be serialized (e.g., .pth) or converted to match the input format of the target 3D deep learning framework.

**Considerations for Dataset Use**

*Intensity characteristics*

The recorded intensity values should be interpreted as relative measurements influenced by scanner position and environmental conditions during acquisition. They are affected by multiple factors, including distance from the scanner and environmental conditions such as weather, surface moisture, and dust accumulation.

In construction environments, different materials produce distinct LiDAR return responses. Even within the same semantic class, reflectance may vary with construction phase or surface condition; for example, a wall may appear as exposed formwork, freshly cast concrete, or gypsum board installation. Consequently, intensity should not be interpreted as an intrinsic material identifier independent of geometric context.

In practice, note that each scan is stored in a local coordinate system centered at the scanner origin. Preprocessing steps that fragment a scan and normalize each fragment independently may distort the original scanner-relative distance patterns encoded in both point density and intensity. Users should therefore apply coordinate normalization with care when preparing fragmented inputs for learning pipelines.

*Class imbalance and long-tail effects*

SIP preserves realistic single-scan construction scenes in which many objects occupy only a small fraction of points within a scan. Slender construction elements such as pipes, ladders, and temporary assemblies may appear with very limited observations, and some objects occur in only a small number of scans. These characteristics naturally produce data scarcity and long-tailed class distributions. Such imbalance affects 3D model training: structural classes with large spatial extent dominate the data, while slender construction objects receive far fewer samples. As a result, segmentation performance for rare classes may degrade substantially and, in extreme cases, remain close to zero during training.

For stable baseline experiments, we recommend using a subset of consistently observed classes in the SIP-Indoor configuration, including wall, ceiling, floor, pipes, ladder, and stair. This recommendation reflects practical training stability given the limited observations of certain categories, while the remaining classes can serve as test cases for long-tail learning and rare object recognition.

Addressing class imbalance remains an open challenge in construction-oriented 3D perception, and the continued release of public construction datasets will be essential for advancing learning methods under real jobsite conditions.

## CONCLUSION

The SIP dataset provides construction-phase TLS scans with realistic occlusion, incomplete visibility,

and rich annotations for semantic segmentation and scene understanding. By releasing SIP and its supporting metadata and utilities, we aim to support research in construction automation, robust 3D perception, and benchmarking under practical jobsite conditions. Future extensions will include expanded scenes and accompanying model benchmarks.

## DATA AVAILABILITY STATEMENT

The SIP dataset introduced in this paper is openly available in a permanent, DOI-assigned repository. The full dataset, including raw LiDAR scans, point-level annotations, and metadata files, can be accessed at: https://doi.org/10.5281/zenodo.17667735. The dataset is released under the Creative Commons Attribution-NonCommercial 4.0 (CC BY-NC 4.0) license. The non-commercial clause reflects institutional considerations related to data collection at the Georgia Tech Student Center construction project, while still allowing free access and reuse for academic, educational, and non-profit research.